\documentclass[conference]{IEEEtran}
\IEEEoverridecommandlockouts
\usepackage{cite}
\usepackage{amsmath,amssymb,amsfonts}
\usepackage{algorithmic}
\usepackage{subcaption}
\usepackage{graphicx}
\usepackage[linesnumbered,ruled,vlined]{algorithm2e}
\usepackage[shortcuts]{extdash}
\usepackage{graphicx}
\usepackage{textcomp}
\usepackage{xcolor}
\def\BibTeX{{\rm B\kern-.05em{\sc i\kern-.025em b}\kern-.08em
    T\kern-.1667em\lower.7ex\hbox{E}\kern-.125emX}}
\begin{document}

% \title{Inferring Implicit Trait Preferences from Demonstrations of Task Allocation in Heterogeneous Teams}
\title{Inferring Implicit Trait Preferences for Task Allocation in Heterogeneous Teams}

\author{\IEEEauthorblockN{Vivek Mallampati,Harish Ravichandar}
\IEEEauthorblockA{\textit{Georgia Institute of Technology} \\
Atlanta, USA \\
\{vmallampati6,harish.ravichandar\}@gatech.edu}
% \and
% \IEEEauthorblockN{Harish Ravichandar}
% \IEEEauthorblockA{\textit{Georgia Institute of Technology} \\
% Atlanta, USA \\
% harish.ravichandar@gatech.edu}
}

\maketitle

\begin{abstract}
Task allocation in heterogeneous multi-agent teams often requires reasoning about multi-dimensional agent traits (i.e., capabilities) and the demands placed on them by tasks. However, existing methods tend to ignore the fact that not all traits equally contribute to a given task. Ignoring such inherent preferences or relative importance can lead to unintended sub-optimal allocations of limited agent resources that do not necessarily contribute to task success. Further, reasoning over a large number of traits can incur a hefty computational burden. To alleviate these concerns, we propose an algorithm to infer task-specific trait preferences implicit in expert demonstrations. We leverage the insight that the consistency with which an expert allocates a trait to a task across demonstrations reflects the trait's importance to that task. Inspired by findings in psychology, we account for the fact that the inherent diversity of a trait in the dataset influences the dataset's informativeness and, thereby, the extent of the inferred preference or the lack thereof. Through detailed numerical simulations and evaluations of a publicly-available soccer dataset (\texttt{FIFA 20}), we demonstrate that we can successfully infer implicit trait preferences and that accounting for the inferred preferences leads to more computationally efficient and effective task allocation, compared to a baseline approach that treats all traits equally.
\end{abstract}

% \begin{IEEEkeywords}
% Task allocation,
% Coalition formation,
% Preference learning, and
% Learning from Demonstrations

% \end{IEEEkeywords}

\section{Introduction}

% Heterogeneous MAS and MRTA are useful and well-studied. 
Heterogeneous multi-agent systems have demonstrated their effectiveness in solving complex problems in a wide variety of domains, such as agriculture~\cite{Tokekar2013Agriculture}, environmental monitoring~\cite{Salam21,Micael19}, defense~\cite{Cook2007Defense}, construction~\cite{Werfel2014Construction},  warehouse automation~\cite{Baras2019Warehouse}, and assembly~\cite{stroupe2005Assembly}. To effectively coordinate such heterogeneous teams, researchers have been studying the Multi-Robot Task Allocation (MRTA) problem~\cite{gerkey_formal_2004} and have developed various solution frameworks, ranging from market-based approaches that auction tasks to agents based on their utilities~\cite{ayorkor_comprehensive_2013}, to the recent advances in trait-based approaches that reason about how agents' capabilities come together to satisfy multi-dimensional requirements placed on them by tasks~\cite{ravichandar_strata_2020,messing_grstaps_2022}. 

%Existing MRTA solutions don't address the problem of preference learning for the traits. 
However, existing approaches to MRTA often consider all traits (i.e., capabilities) to be equally beneficial to a task, ignoring underlying human preferences or the relative importance between traits\cite{ravichandar_strata_2020,messing_grstaps_2022}. For instance, while the color of a robot might be crucial for a task that requires camouflage, it is likely to be irrelevant in most other tasks. Ignorance of such distinctions can lead to sub-optimal task allocations that attempt to satisfy irrelevant requirements, placing an additional and unnecessary burden on the team's limited resources (i.e., agents and their capabilities).

% We model and infer trait preferences and it is useful to do so
In this work, we model and infer task-specific preferences over traits in contexts requiring the satisfaction of specific multi-dimensional requirements. While preference learning is a well-studied problem in other fields (e.g., Human-Robot Interaction (HRI)~\cite{HRI_survey}), 
it is often overlooked in MRTA problems. We argue and demonstrate that inferring and explicitly accounting for such preferences will lead to a more effective allocation of agents to tasks, especially when dealing with teams with limited capabilities and tasks that demand the satisfaction of certain minimum requirements.

\begin{figure}
    \centering
    \includegraphics[width=0.9\columnwidth]{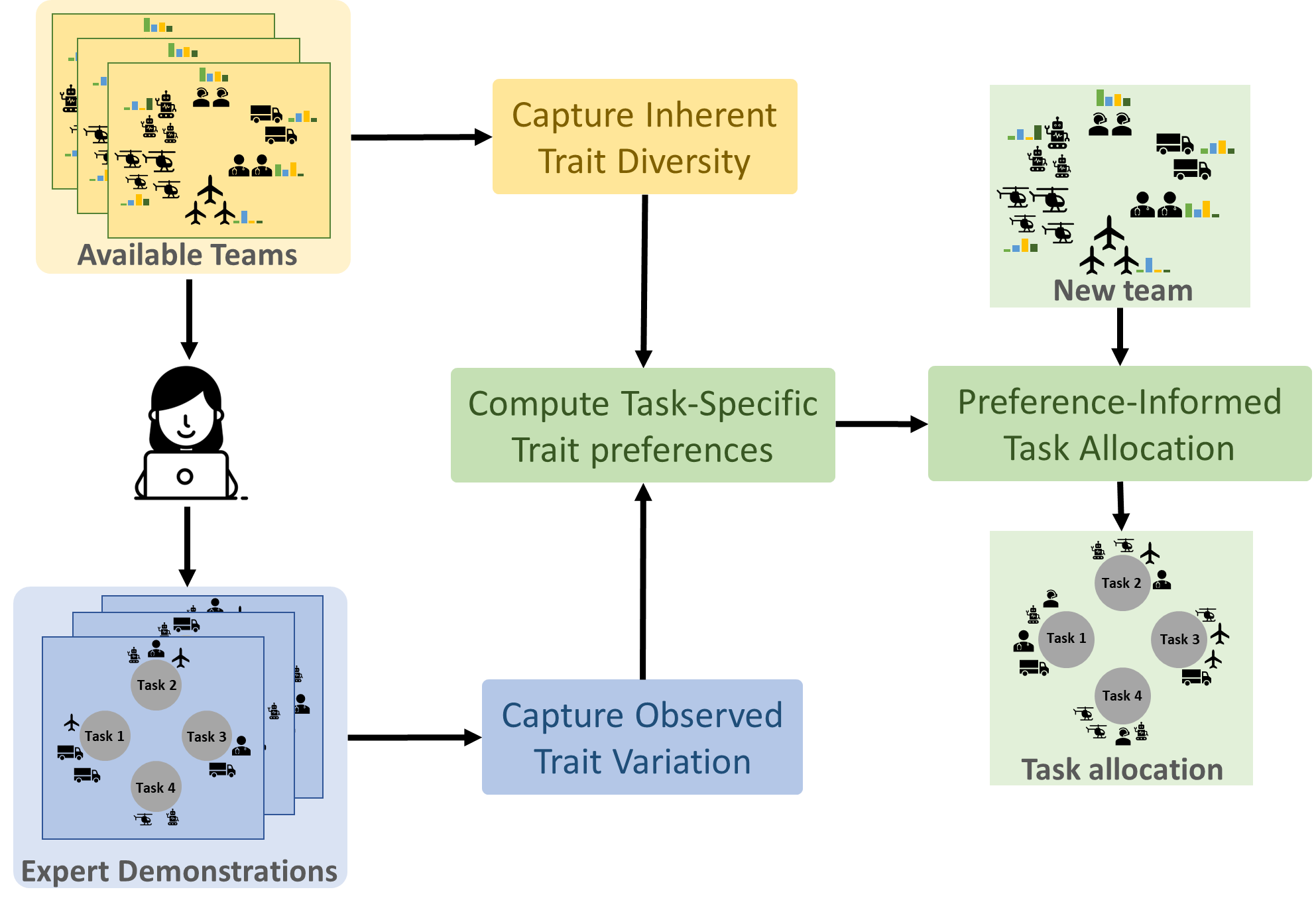}
    \caption{\small{An overview of the proposed approach to infer implicit trait preferences from expert demonstrations and perform coalition formation based on the inferred preferences.}}
    \label{fig:pipeline}
\end{figure}

%We infer the implicit preferences from the expert demonstrations, which will be harder for humans to mention explicitly.
One way to account for trait preferences is to require users to specify them. However, it is known that humans struggle to \textit{explicitly} specific preferences that capture trade-offs between various factors but are adept at solving complex problems with such characteristics~\cite{rieskamp2003people}. As such, we take the approach of inferring preferences that are \textit{implicit} in expert demonstrations of task allocation.

%The importance of a trait is inversely proportional to the variety of the allocations. 
To infer implicit preferences, we first leverage the insight from psychology\cite{sampo_1988_relevance} that the more important or preferred a trait, the more consistent its appearance in the demonstrations (i.e., \textit{observed variation}). For example, if a task is consistently allocated agents who can move at a certain speed, speed is likely an important trait for that task. We compute the coefficient of variation of each trait assigned to a task to capture its consistency. 

%Capturing consistency alone is insufficient. %Using psychology studies about inferring preferences, we consider the abundance of resources available for that particular trait in calculating the importance. 
While capturing the consistency of a trait helps infer preference, it is often insufficient. Specifically, the extent to which a trait can appear to be consistent in the demonstrations is limited by how variable the corresponding capability is in the team prior to allocation (i.e., its \textit{inherent diversity}). Indeed, studies in psychology literature demonstrate that humans' ability to infer preferences increases with the diversity of demonstrations~\cite{kushnir_young_2010}. For instance, if all teams used in the demonstrations only have agents that move at the same speed, the consistency noted in the above example would be inevitable and uninformative. To account for this influence, we develop a method to combine observed variation with inherent diversity such that inherent diversity limits the range of possible preference values based on the teams that were available to the expert, and observed variation informs the specific value of importance within this range based on the expert demonstrations.

% We perform preferential task allocation based on inferred preferences
Finally, we formulate a constrained optimization problem that allocates agents to tasks while accounting for the inferred trait preferences. Notably, since the inferred preferences are over traits (and not robots), we can perform preference-informed task allocation on teams with entirely new agents as long as we have access to the agents' traits. Further, we can improve computational efficiency by reducing of the number of traits the optimization problem needs to consider by removing traits with preference values below a certain threshold. Figure \ref{fig:pipeline} provides a complete overview of our approach.

In summary, our contributions include
\begin{itemize}
    \item A method to infer implicit trait preferences from expert demonstrations by observing the consistency (or variation) in allocations,
    \item Incorporating insights from psychology to alter inferred preferences based on the inherent diversity of each trait in the teams prior to allocation, and
    \item A constrained optimization routine capable of forming coalitions based on the inferred preference.
\end{itemize}

%Our approach shows better performance in metrics like computational efficiency and allocation quality when tested on both simulations and publicly available datasets.  
We evaluated our approach using extensive numerical simulations and a publicly-available soccer dataset (\texttt{FIFA 2020}) and compared its performance against that of a baseline that treats all traits equally. Our results conclusively demonstrate that inferring and accounting for trait preferences i) improves computational efficiency and allocation quality and ii) can significantly reduce the number of dimensions (i.e., traits) that need to be considered with little to no drop in allocation quality.

%%%%%%%%%%%%%%%%%%%%%%%%%%%%%%%%%%%%%%%%%%%%%%%%%%%%%%%%%%%%%%%%%%%%%%%%
\section{Related Works}

Reasoning on how to infer preferences from an expert demonstration and use them in task allocation is a combination of several different research fields. The total problem can be split into two main sub-problems: i) inferring preferences and ii) task allocation. 
Even though there is extensive work done in each of the fields, the lack of literature on the intersection has caught our attention and motivated us to take up this problem. We believe that leveraging the insights in each of field and combining them will provide us with better performance of heterogeneous teams.

In this section, we discuss the prior works done on the topics which are closer to our problem and have solved a part of the problem we are looking into. 

\subsection{Coalition Formation}
According to the survey~\cite{kushnir_young_2010} of the field of MRTA, the problems can be categorized based on three dimensions: robot type (Single Robot (SR) vs. Multi-Robot (MR)), task type (Single-Task (ST) vs. Multi-Task(MT) robots), and Allocation type (Instantaneous assignment (IA) vs. time-extended assignment (TA)). Our work addresses the coalition formation problem, which is an instance of the ST-MR-IA. %Refer to the latest survey~\cite{ayorkor_comprehensive_2013} for a more detailed categorizations of the field. 
% Acknowledging that problem of multi-agent planning and scheduling are close to coalition formation, the fundamental assumption of having access to the action models don't hold in the problem of coalition formation.

Coalition formation (ST-MR-IA) problems have been tackled with utility-based, market-based approaches~\cite{ayorkor_comprehensive_2013}, and trait-based~\cite{ravichandar_strata_2020} approaches. Utility-based approaches showed that they could form optimal coalitions~\cite{vig_2006_utility_coalition}, but designing the utility functions requires domain knowledge and often for specific tasks and agents. The market-based approaches have successfully tackled a variety of coalition formation problems using auction mechanisms through which robots bid on tasks~\cite{guerrero2003auction}. The extensive reliance on the communication between the agents and bidding tasks limits the approach's scalability. The trait-based task allocation models the agents with capabilities (i.e., traits) and tasks as trait aggregations. The framework provides with the advantages of scalability of the size of the robot team and generalizability to new robot teams~\cite{ravichandar_strata_2020, srikanthan_resourceaware_2021, messing_grstaps_2022}. We adopt this framework and model our agents as a set of traits and the tasks as trait aggregation requirements. All the approaches assume that the task is well-defined, which is not true in complex real-world settings.

\subsection{Learning from Demonstrations}
Trying to tackle the challenge of not knowing the task requirements and all constraints, there are developed learning algorithms that try to learn from the expert human demonstrations~\cite{ panait2005cooperative}. Learning methods are also used for market-based coalition formation approaches where the algorithm was trying to find the prices\cite{duvallet2010imitation}. There has been a structured prediction approach~\cite{carion2019structured} for task allocation that uses a combination of reinforcement learning and quadratic integer programming for learning directly from data to optimize assignments of homogeneous teams of robots. 
The recent literature using the trait-based framework enables us to extract the task requirement and work under resource constraints~\cite{srikanthan_resourceaware_2021}. 
Learning from the demonstrations for trait-based task allocation~\cite{srikanthan_resourceaware_2021} enables the algorithms to learn the task requirements and generalize it to new robot teams. 
The one fundamental assumption the existing trait-based task allocation methods make is that they treat all the traits to be equally important. Our approach tries to loosen this assumption and uses the inferred preferences in the optimization of coalition formation. 

\subsection{Preference Learning}

Preference is a very prevalent topic in the fields relating to human psychology~\cite{kahneman_1982_preference}. From child psychology\cite{kushnir_young_2010} to the field of Human-Robot Interaction (HRI) ~\cite{HRI_survey}, preference has been studied in detail and has been given great importance. Literature in human behavior\cite{sampo_1988_relevance} talks about how "not all dimensions of personality are equally relevant to all persons," which in our work is true with the traits of the agents. Even though it is fascinating, to the best of our knowledge, the concept of preference is ignored in the MRTA approaches and not considered in solving coalition formation problems. 

%%%%%%%%%%%%%%%%%%%%%%%%%%%%%%%%%%%%%%%%%%%%%%%%%%%%%%%%%%%%%%%%%%%%%%%%

\section{Problem formulation}
% This section explains the problem formulations and the definitions. 
In this section, we formulate the problem of inferring trait preferences in heterogeneous task allocation.
We begin with basic terminology and definitions inspired by recent advances in trait-based task allocation~\cite{ravichandar_strata_2020,srikanthan_resourceaware_2021,messing_grstaps_2022}.

Let a heterogeneous team composed of $S$ species (i.e., types of agents) with $A_s$ agents in the $s^{th}$ species. This heterogeneous team is required to perform $M$ concurrent tasks denoted by $T = \{T_1, \dots, T_M \}$.

The traits (i.e., capabilities) of the team are encoded by its \textbf{Species-Trait matrix} 
$Q = [q_1, \dots, q_S]^T \in \mathbb{R}_{\geq 0}^{S\times U}$, where $q_s \in \mathbb{R}_{\geq 0}^U$ is a vector of $U$ number of traits associated with Species $s$.

The \textbf{Assignment matrix} $X \in \mathbb{Z}_{\geq 0}^{M\times S} $ encodes the allocations, where Assignment $x_{ms}$ is the number of agents from Species $s$ assigned to Task $T_m$.

The \textbf{Trait-Requirement matrix} $Y \in  \mathbb{R}_{\geq 0}^{M\times U}$ given by 
\begin{eqnarray}\label{eq:1}
{\mathbf{Y} = \mathbf{X}\mathbf{Q}}
\end{eqnarray}
captures the aggregation of traits assigned to tasks, with $Y_{mu}$ indicating the amount of the Trait $u$ is required by Task $T_m$.

Let $\mathcal{D}= \{X^{(i)}, Q^{(i)}\}^N_{i=1}$ be the set of \textbf{Expert Demonstrations}, where $X^{(i)}$ represents the expert-specified assignment matrix
for a team with traits encoded by the Species-Trait matrix $Q^{(i)}$. We assume that the expert allocates tasks by minimizing the following weighted cost function
\begin{eqnarray}\label{eq:2}
    X^{(i)} =\arg \min_{X} || {Y^*} - X Q_i||_{W^*}, \forall i=1,\cdots,N
\end{eqnarray}
where $Y^*$ and $W^*\in\mathbb{R}_{\geq 0}^{M \times U}$ are hidden trait requirements and the associated trait preferences employed by the expert.

\textbf{Problem Statement}: Given Demonstrations $\mathcal{D}$, we wish to infer implicit trait requirements $\hat{Y}^*$ and the associated trait preferences $\hat{W}$.

%%%%%%%%%%%%%%%%%%%%%%%%%%%%%%%%%%%%%%%%%%%%%%%%%%%%%%%%%%%%%%%%%%%%%%%%
\section{Inferring Trait Preferences}  

Our proposed approach extracts the trait preferences from the given set of demonstrations which will be used to enhance the task-trait assignments.
A secondary objective is to fulfill task requirements using as few traits as possible to reduce the time it takes to optimize the assignment. We begin by computing the aggregated traits across tasks for each demonstration as follows: 
\begin{eqnarray}\label{eq:3}
    Y^{(i)} = X^{(i)}Q^{(i)}, \forall i=1,\cdots,N
\end{eqnarray}

\noindent The following two key factors inform our inference method:
\begin{enumerate}
    \item \textbf{Observed Variation} ($CV_{obs}$) captures the variation of each aggregated trait across the demonstrations for each task. Observed variation captures how consistently a trait is allocated to a particular task. Our insight here is that the importance of a trait to a task is inversely proportional to its variability in the demonstrations.
    Note that the lower the observed variation, the higher the chances that the corresponding trait was deliberately chosen to reflect a consistent value in the allocation. In contrast, larger observed variation tends to stem from lower preferences for the associated trait. The key idea is that a trait that a wide variety of values can fulfil is relatively less preferable as the expert did not focus on matching a specific value to the considered trait.  The observed variation is calculated with the equation
    \begin{equation}\label{eq:4}
    {CV_{obs} = \frac{\sigma(Y_{\mathcal{D}})}{\overline{Y}_{\mathcal{D}}}}
    \end{equation}
    where $Y_{\mathcal{D}} \in \mathbb{R}^{N \times M \times U}$ is the collection of all aggregated trait matrices from the demonstrations ${Y}^{(i)}, \forall i=1,..., N$, 
    
    $\sigma(Y_{\mathcal{D}})$ is the standard deviation of $Y_{\mathcal{D}}$, and $\overline{Y}_{\mathcal{D}}$ is the mean of $Y_{\mathcal{D}}$, both computed across demonstrations.
    
    \item \textbf{Inherent Diversity} ($CV_{div}$) captures the diversity of each trait in the demonstrations $\mathcal{D}$. Note that the more abundant a trait, the more informative the dataset. If all agents in the dataset possess near-identical amounts of a particular trait, we cannot rely solely on the inevitable low observed variation and conclude that the trait in question is highly preferred. The inherent diversity is calculated with the equation \begin{equation}\label{eq:5}
    {CV_{div} = \frac{\sigma(Q_{\mathcal{D}})}{\overline{Q}_{\mathcal{D}}}}\end{equation}
    where $Q_{\mathcal{D}} \in \mathbb{R}^{(N*S) \times U}$ is the collection of all species-trait matrices from the demonstrations ${Q}^{(i)}, \forall i=1,...,N$, 
    
    $\sigma(Q_{\mathcal{D}})$ is the standard deviation of $Q_{\mathcal{D}}$, and $\overline{Q}_{\mathcal{D}}$ is the mean of $Q_{\mathcal{D}}$.
    
\end{enumerate}

Note that we use the coefficient of variation and not the variance of the aggregations, as it permits scale-independent comparisons~\cite{Brown1998} while quantifying the "spread" of the data. Directly computing variances will mislead the algorithm to assign higher importance to lower-magnitude traits than higher-magnitude traits. 

In humans, inferring preferences is a very natural phenomenon \cite{kushnir_young_2010}. Psychology studies on human behavior explain how consistent selection and human preference are related ~\cite{sampo_1988_relevance}. This insight helps us in modeling the Observed Variation as a variable that helps infer relative importance.  We also came across a child psychology study on inferring preferences that indicates a linear relationship between to variation of the demonstrations and the average number of correct target picks (most preferred object) \cite{kushnir_young_2010}. We try to incorporate this idea of the importance of a variety of trait values in demonstrations through the concept of Inherent Diversity.

\textbf{Weight Function}: We propose the following weight function that encodes the preference over traits by combining Observed Variation and Inherent Diversity. 

\begin{equation}\label{eq:6}
    {\hat{W} = \frac{CV_{div}}{\tau}cos(\alpha CV_{obs} + \beta) + \textit{c}}
\end{equation}

Note that the weight function is a cosine function whose amplification factor is dictated by Inherent Diversity, and the independent variable is Observed variation. We used the cosine function to combine because of the linear amplification feature, and with scaling and translations, it is monotonically decreasing in the desired range. $\tau $ in the equation \ref{eq:6} is the amplification factor. It regulates how much the inherent diversity affects the weight function. $\alpha$ and $\beta$ are the factors that manipulate the observed variation and its impact on the weight function. The range of the weight function and the possible output values can be altered by the combination of $\alpha$ and $\beta$. Lastly, $\textit{c}$ is the centering variable. The weight function output equal to $\textit{c}$ means that it essentially has no information about the relative preference. 

\begin{equation}\label{eq:6a}
    {\hat{W} = \frac{CV_{div}}{2}cos(2 CV_{obs} + 0.5) + 0.5}
\end{equation}

\begin{figure}[ht]
\centering
\includegraphics[width=0.75\columnwidth]{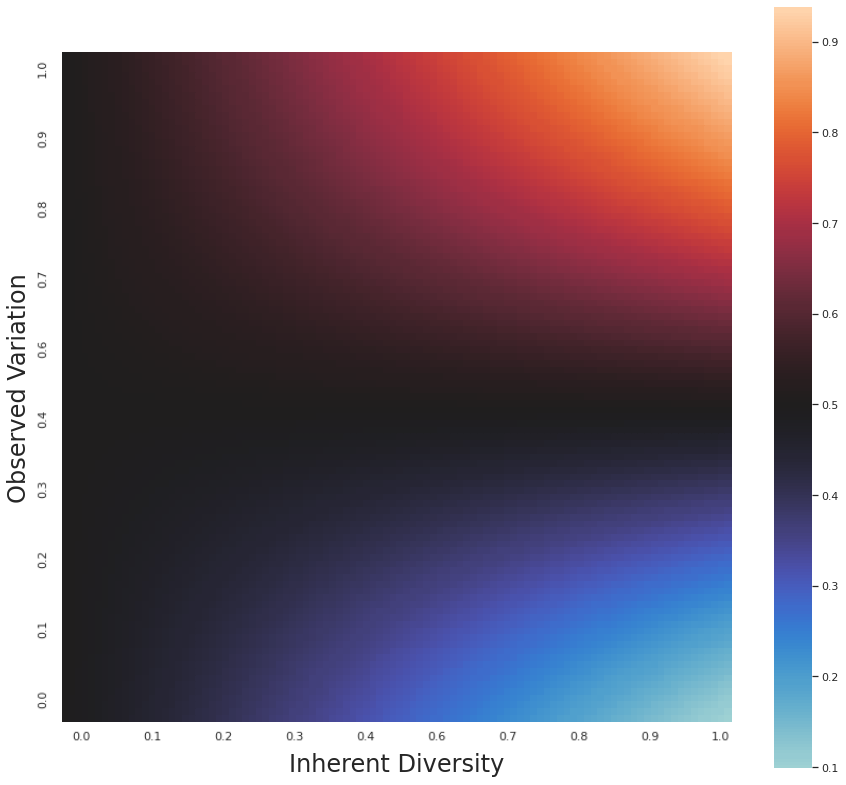}
\caption{Heatmap of the outputs of the proposed weight function}
\label{fig:weight-heatmap}
\end{figure}

We have defined with $\alpha = 2$,$\tau = 2$, and $\beta = 0.5$ to make the inherent diversity and observed variation in the range [0,1] and have the inferred weights values be in the range [0,1]. The weight function in Eq \ref{eq:6a} value of 0 means it is not at all important, and the value of 1 means it is the most important. The result of the weight function being 0.5 indicates that the trait preference is inconclusive. The $\textit{c}=0.5$ in the weight function helps us to show that having no inherent diversity will result in inconclusive weight preference as we can not interpret the lower observed variation because of a lack of choice in the dataset or because it was an extremely important trait value. The heat-map from Figure \ref{fig:weight-heatmap} displays the values of the equation \ref{eq:6a} over the total input space. We can clearly see that as the inherent diversity increases, the observed variance's possible values can increase. Inherent diversity acts like a range-increasing function for the observed variation variable in the weight function. 

The algorithm for the weight extraction can be seen in Algorithm \ref{alg:weight-extraction} to see the steps for extracting variations and finding the weight matrix for the given set of demonstrations. 

% \begin{algorithm}
%     \caption{Inferring Trait Preferences}
%         \label{alg:weight-extraction} 
%     \begin{algorithmic}
%         \Require Demonstrations $\mathcal{D} = \{X^{(i)}, Q^{(i)}\}_{i=1}^N$
%         \State $Y_{\mathcal{D}} \gets [X^{(i)}*Q^{(i)}],  \forall i=1,...,N$   
%         \State $Q_{\mathcal{D}} \gets$ \{ aggregate$(Q^{(i)}) \}, \forall i =1,...,N$
%         \State $CV_{div} \gets$ coefficient of variation$(Q_{\mathcal{D}})$
%         \State $CV_{obs} \gets$ coefficient of variation$(Y_{\mathcal{D}})$
%         \State $\hat{W} \gets  \frac{CV_{div}}{2} cos(2CV_{obs} + 0.5) + 0.5$
%         \State return $\hat{W}$
%     \end{algorithmic}
% \end{algorithm}

\begin{algorithm}
  \SetNoFillComment   
  \SetKwInOut{Input}{Input}
  \SetKwInOut{Output}{Output}
  \SetKwProg{try}{try}{:}{}
  \SetKwProg{catch}{except}{:}{end}
  \Input{Demonstrations $\mathcal{D} = \{X^{(i)}, Q^{(i)}\}_{i=1}^N$}
  \Output{$\hat{W}$} 
Calculate the trait-aggregation $Y_{\mathcal{D}}$ as shown in Equation \ref{eq:3}
\\Aggregate all the trait from demonstrations into $Q_{\mathcal{D}}$
\\Calculate the observed variation $CV_{obs}$ as shown in Equation \ref{eq:4}
\\Calculate the inherent diversity $CV_{div}$ as shown in Equation \ref{eq:5}
\\Calculate the inferred weight $\hat{W}$ using the Equation \ref{eq:6a}
    
\Return $\hat{W}$
  \caption{Inferring Trait Preferences}
        \label{alg:weight-extraction} 
\end{algorithm}

The algorithm for the task allocation, unlike the optimizations in prior works~\cite{ravichandar_strata_2020,messing_grstaps_2022,srikanthan_resourceaware_2021}, incorporates weights. One critical note is that the estimation of $\hat{Y}^*$ is done simply by taking the mean of the $Y_{\mathcal{D}}$ we calculated from the demonstrations. The key assumption here is that all the demonstrations are optimal; hence, the average of the task-trait aggregations will provide us with the optimal $\hat{Y}^*$ that the new teams need to reach. The algorithm \ref{alg:task-allocation} outlines the procedure taken by the algorithm to do the weighted optimization for allocating a new team of agents for the same set of tasks. 

% \begin{algorithm}
%     \caption{Preferential Task Allocation}
%         \label{alg:task-allocation} 
%     \begin{algorithmic}
%         \Require $Q_j$, $\hat{W}$, $Y_{\mathcal{D}}$ where $Q_j$ is the new team. $\hat{W}$ is the inferred weight. $Y_{\mathcal{D}}$ is the collection of all the $N$ task-trait matrices.       
%         \State $\hat{Y}^* \gets$ mean of $Y_{\mathcal{D}}$ 
%         \State $\hat{X}_j =\underset{X}{\mathrm{argmin}} || {\hat{Y}^*} - X Q_j||_{\hat{W}}$
%         \State return $\hat{X}_j$
%     \end{algorithmic}
% \end{algorithm}

\begin{algorithm}
  \SetNoFillComment   
  \SetKwInOut{Input}{Input}
  \SetKwInOut{Output}{Output}
  \SetKwProg{try}{try}{:}{}
  \SetKwProg{catch}{except}{:}{end}
  \Input{New species-trait matrix ($Q_j$), $\hat{W}$, and $Y_{\mathcal{D}}$}
  \Output{$\hat{X}_j$} 
    $\hat{Y}^* = $ mean of $Y_{\mathcal{D}}$
    \\$\hat{X}_j =\underset{X}{\mathrm{argmin}} || {\hat{Y}^*} - X Q_j||_{\hat{W}}$
    
  \Return $\hat{X}_j$
  \caption{Preferential Task Allocation}
        \label{alg:task-allocation} 
\end{algorithm}

\section{Experimental Evaluation}

We evaluated the efficacy and need for the proposed approach using the following sets of experiments.

\subsection{Numerical Simulations}
\noindent\textit{Setup}: The first step of the experiment is creating a demonstration set that mimics expert behavior. The basic setup of the demonstrations included agents from three species allocated to three tasks. Each of the species has three traits and 15 agents available. 
Handcrafting the optimal weights ($W^*$), which describe the behavior of the demonstrations, we created a pipeline that consistently models Species-trait matrices $Q$ and solves for the assignment matrices $X$ using equation \ref{eq:2} to simulate the expert demonstrations for allocating robots for the given tasks. 
In order to simulate the $Q$ for the demonstrations, we use the following hand-crafted mean ($\mu$) and standard deviation ($\sigma$) matrices:

$\mu_{(M\ X\ U)} = 
\begin{bmatrix}
10 & 5 & 20\\
5 & 20 & 10\\
20 & 10 & 5\\
\end{bmatrix}  
\sigma_{(M\ X\ U)} = 
\begin{bmatrix}
2 & 1 & 3\\
1 & 3 & 2\\
3 & 2 & 1\\
\end{bmatrix}$

To make the simulations extensive, we create 1000 demonstrations with various optimal weights which simulate intrinsic preferences. 
Across all the three tasks, few optimal weights we test on are:
\begin{enumerate}
    \item Single important trait: $W^* = [0,1,0]$ This is where there is one most crucial trait and model the demonstration such that the expert only considers that trait when allocating. 
    \item Unique Large Weight differences: $W^* = [0.6,0.3,0.1]$ This is where there is one trait that has a lot higher preference. The three traits are non zero distinct values indicating that the user has some preference associated.
    \item Equal Large Weight differences: $W^* = [0.8,0.1,0.1]$ This is where there is one trait that has a lot higher preference. The other two traits are equal non zero indicating that the user has some preference associated.
    \item Closer Weights: $W^* = [0.5,0.3,0.2]$ This is where all three traits have some preference. We can see that relative importance is different for each trait. 
\end{enumerate}

\noindent\textit{Baselines}:
We compare the performance of our approach against the following baselines:
\begin{enumerate}
    \item \textbf{No Preference}: The baseline was implemented by previous works\cite{ravichandar_strata_2020, srikanthan_resourceaware_2021} where the preferences of the traits are not considered. The baseline uses the $\hat{Y}^*$ shown in the algorithm \ref{alg:task-allocation}. Calculates the $\hat{X}_j$ using the equation \ref{eq:2}
    
    \item \textbf{No Inherent Variation}: The baseline follows the algorithm \ref{alg:task-allocation}, with a variation in algorithm \ref{alg:weight-extraction}. The baseline skips line 4 of the algorithm \ref{alg:weight-extraction} and assigns $CV_{div}$ to be an array of $0.5$s. Inherent diversity set to 0.5 for all traits in every task indicates that there is no consideration of inherent variation. 
\end{enumerate}

\noindent\textit{Results}: We evaluate our numerical experiments using quantitative metrics on the performance of the task allocation and inferring preferences. 

The primary metric we used to evaluate the inferring of preferences is ordering. 
The ordering of the weights is a crucial metric when we know the optimal weights, 
as the order of preferences is directly implied by the arrangement of the weights.

We observe that for all the different scenarios mentioned in setup, where demonstrations were constructed with different optimal weights, the Algorithm \ref{alg:weight-extraction} 100\% successfully extracts the weights, preserving the weights' ordering across the tasks. Even in the situations where the weights are close, the ordering of the weights were not compromised.

Knowing that the weights inferred are evaluated to be good, we explore the impact of using the weights in task allocation. We use the metric of allocation quality by computing the trait mismatch error. A lower weighted mismatch error indicates that the allocation quality has increased, implying the most important traits have been fulfilled. 

\begin{figure}
\centering
\begin{subfigure}{0.8\columnwidth}
    \includegraphics[width=\columnwidth]{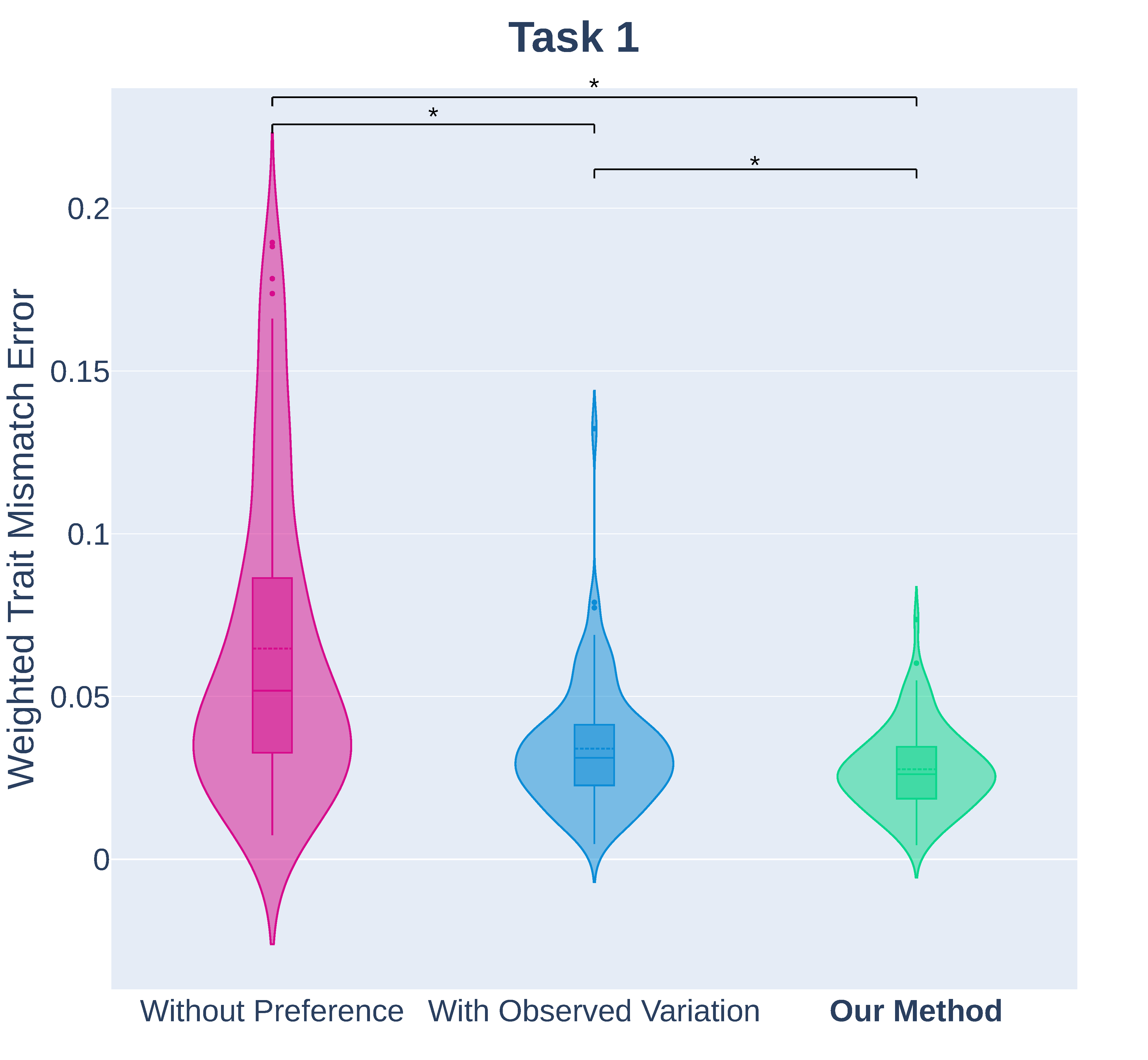}
    \label{fig:task_one}
\end{subfigure}
\begin{subfigure}{0.8\columnwidth}
    \includegraphics[width=\columnwidth]{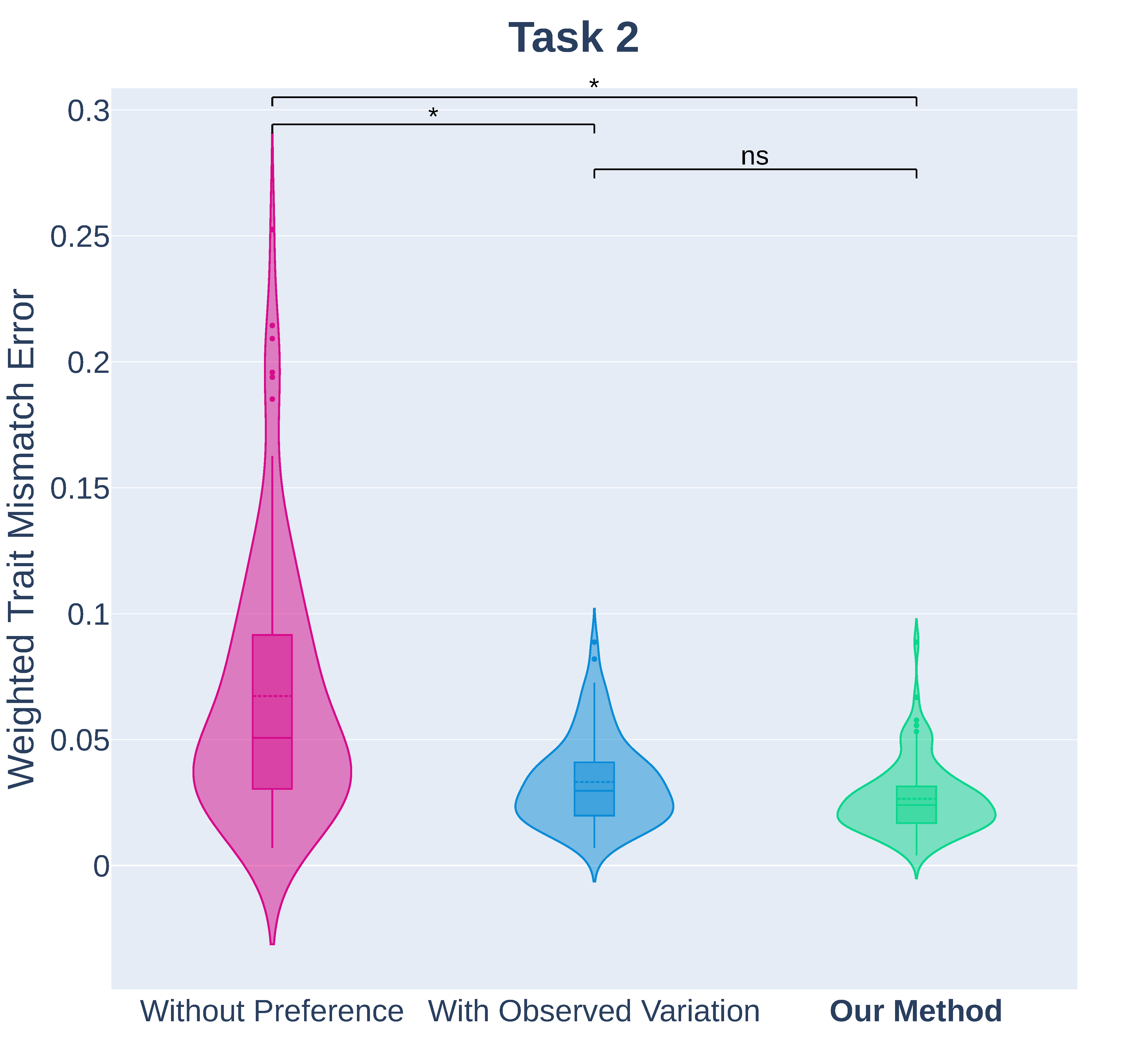}
    \label{fig:task_two}
\end{subfigure}
\begin{subfigure}{0.8\columnwidth}
    \includegraphics[width=\columnwidth]{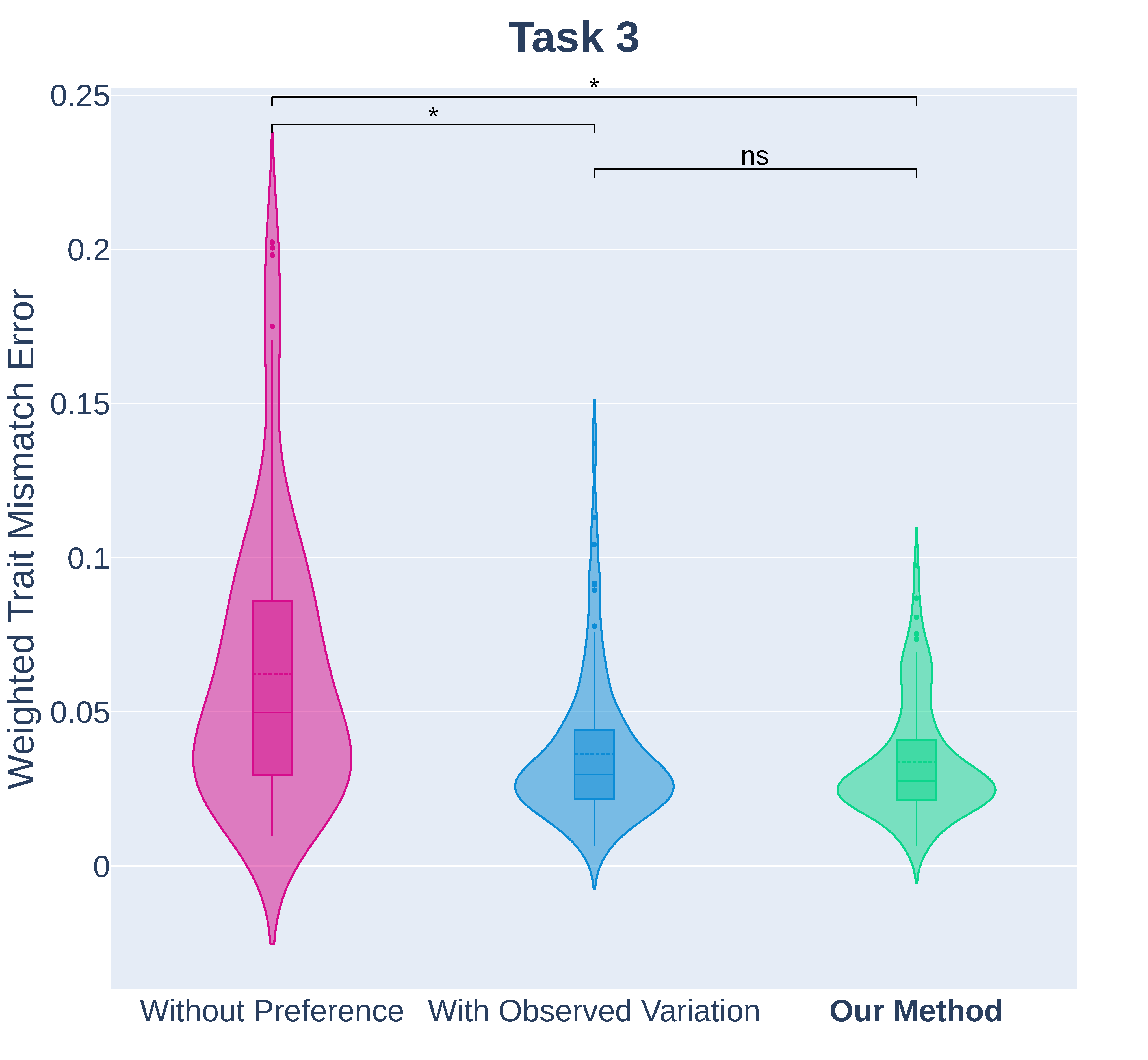}
    \label{fig:task_third}
\end{subfigure}

\caption{In numerical simulations, inferring and accounting for preferences consistently results in lower ground-truth-weighted trait mismatch errors.}
\label{fig:numerical-results}
\end{figure}

Figure \ref{fig:numerical-results} shows the comparison between our approach against the baselines across all three tasks.  
We can observe that across the tasks, our method outperforms the baseline methods by having a median, mode, and mean lower for the weighted error. These observations are consistent across various scenarios with different hand-crafted weights ($W^*$). 

We have done a statistical analysis of the results from all three methods. We ran Kruskal Wallis on all three methods for each task to see that there is significance between all three methods. We went on to do a pairwise Mann-Whitney U Test to check the relationships between the two methods.

 The table below provides the p-value of the Mann-Whitney U Test, in which Pair 1 is the comparison between the two baselines (No preference and No inherent diversity). Pair 2 is the comparison between the No preference baseline and our approach. Pair 3 is the comparison between the No inherent diversity baseline and our approach. 
\begin{table}[ht]
    \centering
    \begin{tabular}{c|c|c|c}
        Task Number & Pair 1 & Pair 2 & Pair 3\\ \hline
        Task 1 & 7.595e-07 & 8.326e-11 & 0.027\\
        Task 2 & 4.388e-08 & 5.061e-12 & 0.057\\
        Task 3 & 1.831e-05 & 1.077e-07 & 0.231\\ \hline
    \end{tabular}
    \caption{Non-parametric test p-values}
    \label{tab:my_label}
\end{table}

Figure \ref{fig:numerical-results} has pair-wise relationships, and in situations where the p-value $>$ 0.05 is denoted with `ns' to indicate non-significance. 
We see a significant difference between the no preference baseline and our approach, indicating that utilizing the preferences results in better trait satisfaction. We can also observe that using only the Observed Variation can enhance performance. Further, using the inherent diversity along with observed variation results in the best allocations in terms of trait satisfaction.

We also investigated if the number of demonstrations plays a crucial role in the extraction of weights and in the weighted error difference between the target requirements $\hat{Y}^*$ and the trait aggregation achieved by the formed coalition ($Y=XQ$). Our results suggest that our ability to infer trait preference gracefully degrades as we reduce the number of demonstrations (1000 demos: 100\%, 900 demos: 98.5\%, 800 demos: 95.4\% … 200 demos: 86.3\%, 100 demos: 81.3\%)

\subsection{Soccer Dataset}

\noindent\textit{Idea}: Most multi-agent task allocation approaches to showcase their results on simulated data or data explicitly created for a set of robots. We have taken a step forward in extending our work to test real-life datasets from the domain of sports. We have chosen the game of \textit{soccer} and tried to map it into a multi-agent task allocation problem. 

Each game position is modeled as the tasks and the players as agents. The task allocation problem, in this scenario, is allocating given agents (players) to one of the four tasks (game positions): forward, midfield, defense, and goalkeeper. 
Given new soccer players with traits, the optimization tries to match the trait aggregation requirement for every task (game position) to create a soccer team of 11 players with 1 goalkeeper, 2 forwards, 4 defenders and 4 midfielders. 
% \newline

\begin{figure}
\centering
\includegraphics[width=0.8\columnwidth, trim={0 0 0 50}, clip]{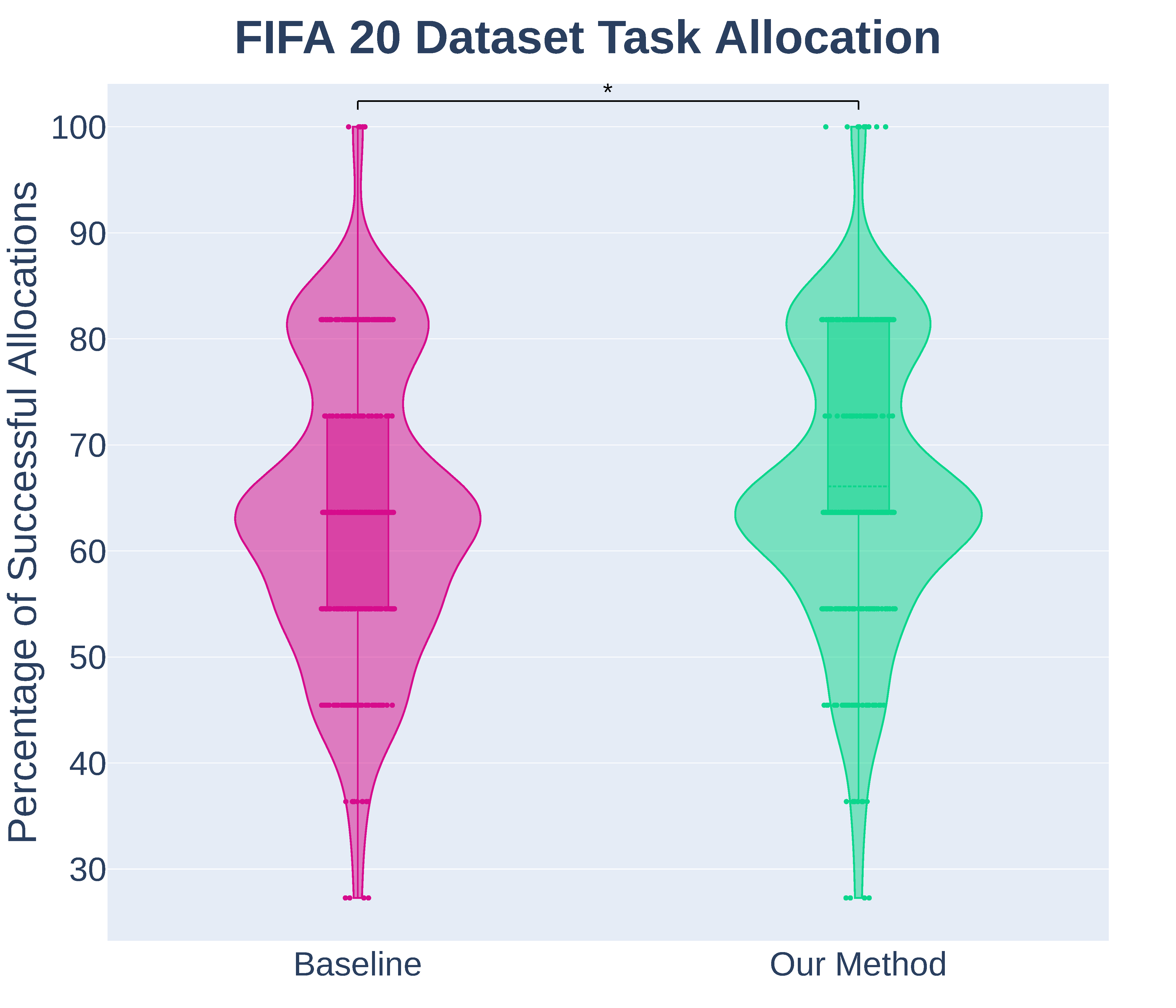}
\caption{Our inference and allocation approach helps significantly increase the percentage of successful allocations in \texttt{FIFA 2020} dataset.}
\label{fig:fifa_comp-results}
\end{figure}

\noindent\textit{Setup}:
Using the open source \texttt{FIFA 2020} dataset from Kaggle~\cite{fifa_data}, we started to design the experiment. Pre-processing the data was crucial to fit the modeling framework we chose in our approach.  
Similar to our numerical simulations, we wanted to have a set of demonstrations. 
The \texttt{FIFA 2020} player dataset has 18278 players, each with 104 columns of information. 
As part of the pre-processing, the dataset was trimmed down only to contain the necessary numerical capabilities of players. 
We removed all the inactive and substitute players and players with missing capabilities to make the dataset consistent. 
To make the demonstrations consistent, we have filtered our dataset to have players with \texttt{FIFA 2020} players' 'Overall' between 60-70 (inclusive). 
The final demonstrations we use have 37 traits and around 3600 players.    
We have taken the average of all players' traits assigned to a particular task (game position) given in the dataset to create the $\hat{Y}^*$. 

From the processed dataset, both the $Q_{\mathcal{D}}$ and $Y_{\mathcal{D}}$ for the soccer experiment were generated directly. 
Knowing the $Q_{\mathcal{D}}$ and $Y_{\mathcal{D}}$, we can use the algorithm \ref{alg:weight-extraction} to get the weight matrix which indicates inferred preferences of traits for every task. 
Creating a new 11-player team from the dataset, we ran algorithm \ref{alg:task-allocation} to see if the algorithm could classify the players to their original game positions in the \texttt{FIFA 2020} dataset. 
  
\noindent\textit{Results}: 
Given the lack of ground-truth trait preference ($W^*$) in this dataset, we cannot directly evaluate the accuracy of our inferred preferences. Instead, we extensively analyzed the impact of preference-based task allocation on computation time and allocation score.

Figure \ref{fig:fifa_comp-results} clearly shows that our method using both observed variation and inherent diversity increases the allocation score. The statistical significance (p-value $>$ 0.05) calculated using Kruskal–Wallis test between the Baseline allocation score and Our Method allocation score, clearly shows our method outperforms the baseline.  

\begin{figure}
    \centering
    \includegraphics[width=0.8\columnwidth]{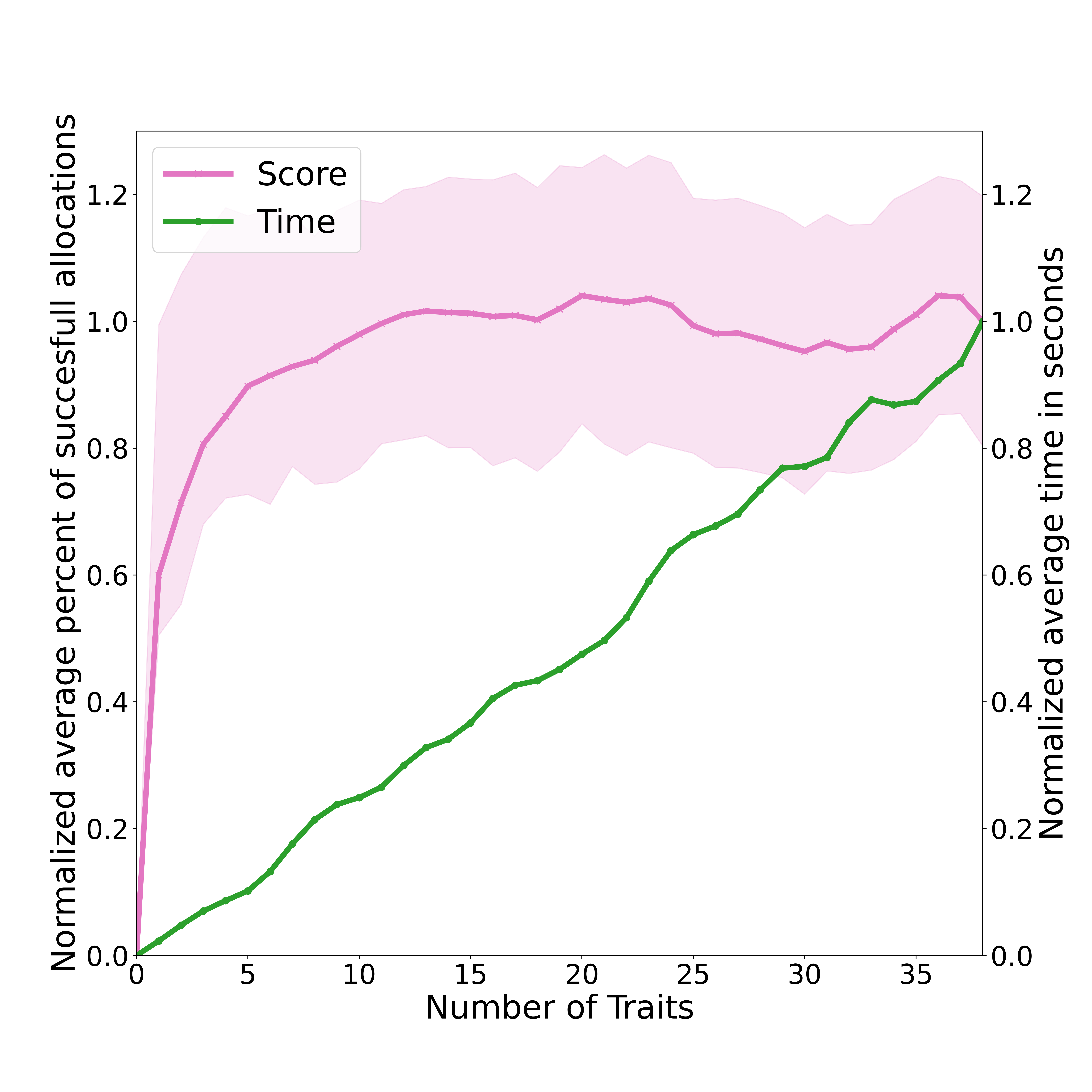}
    \caption{Our inference and allocation approach helps significantly reduce the trait dimensions without sacrificing allocation quality in \texttt{FIFA 2020} dataset.}
    \label{fig:fifa-sim}
\end{figure}

Inferring the preferences has enabled our approach to use the k-most important traits with higher preferences than others, where k is the number of traits we want to use in allocation. Reduction in the dimensionality of the optimization problem has resulted in a decrease in computational time. The green line in Figure \ref{fig:fifa-sim} shows how the time increases with the number of traits used. The time is normalized with the total time taken by the baseline, which doesn't consider preference. 

The second metric of evaluation we used was the successful allocation percentage. We define the successful allocation by comparing the allocations made by the algorithm and the original game positions given in the dataset. The total number of possible allocations is the team size (11 for soccer). The percentage is taken by the number of positions matching between the dataset and allocation divided by team size. The pink line in Figure \ref{fig:fifa-sim} shows how the score (i.e., successful allocations)  increases with the number of traits used. The score is normalized with the score achieved by the baseline, which doesn't consider preference. The light pink region shows the standard deviation of the teams' scores across different traits. 

We have plotted both time and successful allocation percentages across the number of traits to show that our approach can significantly reduce the number of dimensions (i.e., traits) that needs to be considered while allocating agents without compromising allocation quality. We can infer from the graph that an allocation using only 12 trait weights can achieve the baseline's accuracy by being more than 52\% faster. Compromising on less than 5\% of baseline accuracy can result in a reduction of 75\% of the time taken by the baseline. Experiments on the \texttt{FIFA 2020} dataset show that learning preferences and preference-based task allocation can increase task performance.

\section{Conclusion}
We proposed a novel method to infer implicit trait preference from heterogeneous multi-agent task allocation demonstrations. Our method reasons about i) \textit{observed variation}: the consistency with which traits appear across demonstrated allocation, and ii) \textit{inherent diversity}: the variability of traits in the overall teams prior to allocation. Our results demonstrate that inferring and accounting for preferences during allocation can considerably improve trait satisfaction and computational efficiency. While not considered in this work, the proposed preferential task allocation can be integrated into larger frameworks, such as GRSTAPS~\cite{messing_grstaps_2022}, in which task allocation serves as one of the many layers of coordination. While our relatively simple method has proven effective, future work must address how one can reliably infer preferences in more complex settings using advanced learning techniques, such as contrastive learning, especially when dealing with noisy and low-diversity datasets. A notable limitation of our work is that it is limited to contexts with tasks that require meeting specific trait requirements. As such, our method can not handle tasks in which certain capabilities merely need to be maximized. Inferring preferences in such circumstances remains an unsolved problem.

\bibliographystyle{IEEEtran}  
\bibliography{bibtext}

\end{document}